\documentclass[pdflatex,sn-nature]{sn-jnl}

\makeatletter
\@twosidefalse
\makeatother

\usepackage{graphicx}%
\usepackage{multirow}%
\usepackage{amsmath,amssymb,amsfonts}%
\usepackage{amsthm}%
\usepackage{mathrsfs}%
\usepackage[title]{appendix}%
\usepackage{xcolor}%
\usepackage{textcomp}%
\usepackage{manyfoot}%
\usepackage{booktabs}%
\usepackage{algorithm}%
\usepackage{algorithmicx}%
\usepackage{algpseudocode}%
\usepackage{listings}%

\usepackage[left]{lineno}
\usepackage{bookmark}

\theoremstyle{thmstyleone}%
\theoremstyle{thmstyletwo}%

\theoremstyle{thmstylethree}%

\begin{document}

\title{Learning from high-resolution 2D images to reconstruct 3D microstructures}


\author*[1,2]{\fnm{Wenbo} \sur{Sui}}\email{wenbo.sui@thermofisher.com}
\author[3]{\fnm{Lars P.} \sur{Mikkelsen}}
\author[2]{\fnm{Jens W.} \sur{Andreasen}}
\author[1]{\fnm{Daniel} \sur{Lichau}}
\author[1]{\fnm{Harold} \sur{Phelippeau}}
\author[4]{\fnm{Philip J.} \sur{Withers}}

\affil*[1]{\orgname{Thermo Fisher Scientific}, \orgaddress{\city{Bordeaux}, \postcode{33800}, \country{France}}}

\affil[2]{\orgdiv{DTU Energy}, \orgname{Technical University of Denmark}, \orgaddress{\city{Kgs. Lyngby}, \postcode{2800}, \country{Denmark}}}

\affil[3]{\orgdiv{DTU Wind}, \orgname{Technical University of Denmark}, \orgaddress{\city{Roskilde}, \postcode{4000}, \country{Denmark}}}

\affil[4]{\orgdiv{Henry Royce Institute, Department of Materials}, \orgname{University of Manchester}, \orgaddress{\city{Manchester}, \postcode{M13 9PL}, \country{United Kingdom}}}

\abstract{High-resolution three-dimensional (3D) microstructure characterization remains difficult to undertake at scale because volumetric imaging at the required spatial resolution is expensive, time-consuming or destructive. In many settings, however, fine-resolution two-dimensional (2D) imaging and lower-resolution volumetric measurements are both readily available. We introduce a Privileged Cross-modal Relational Distillation (PCRD) framework that leverages fine-resolution 2D observations during training to improve specimen-specific 3D reconstruction from lower-resolution volumetric imaging alone. We demonstrate PCRD on pultruded carbon fiber-reinforced polymer composites, using scanning electron microscopy as the privileged modality and micro-CT as the inference input. The framework transfers spatial organization by matching similarity distributions defined around corresponding spatial anchors in the teacher and student representations, avoiding direct alignment of heterogeneous feature vectors. In 2D cross-sectional reconstruction, PCRD improves mAP by 1.66 percentage points over direct micro-CT training. Independent 3D validation against high-resolution synchrotron X-ray CT data shows close agreement in fiber volume fraction, diameter and local spacing. The PCRD reconstruction matches 89.97\% of the reference fiber centerline length, compared with 75.08\% for the interpolated low-resolution volume.}




\maketitle

\section{Introduction}\label{sec:1}
Three-dimensional (3D) microstructure characterization is fundamental to the study of heterogeneous materials, as effective properties such as transport and mechanical response are governed by the spatial arrangement of constituents, pores and interfaces \cite{greeley2025quantitative, he2025multiscale}. Reliable property computation, physics-based simulation and inverse material design all require 3D image-based modelling based on accurate microstructural data \cite{auenhammer2024x, he2025multiscale, jiang2026ct, saravanan2026assessing}. However, obtaining such data through direct 3D imaging remains expensive, time-consuming or destructive, limiting large-scale dataset construction and routine analysis \cite{salvo20103d, cao2020emerging, mura2023advances}.

Fine-resolution two-dimensional (2D) imaging, including optical microscopy and scanning electron microscopy (SEM), is often considerably easier to undertake than volumetric imaging \cite{he2025multiscale,danalou2026digital}. This accessibility has motivated methods that infer 3D microstructures from representative 2D observations. SliceGAN \cite{kench2021generating} was an early deep generative approach in this direction, training a 3D generative adversarial network from representative 2D micrographs to synthesize volumes with similar structural statistics. Phan et al. \cite{phan2024generating} combined denoising diffusion with adversarial supervision for 3D microstructure generation from a single 2D image, while Micro3Diff \cite{lee2024multi} adapted pretrained 2D diffusion models through coordinated denoising across orthogonal planes. Recent work \cite{wang20263d} has also incorporated explicit statistical descriptors to improve the preservation of structural characteristics. These approaches provide practical routes to statistically representative 3D microstructures when fine-resolution volumetric data are unavailable. However, statistics derived from 2D observations do not uniquely determine the corresponding 3D geometry, and different 3D configurations can satisfy the same constraints. This uncertainty can affect downstream analysis, with SliceGAN-type statistical reconstructions reported to produce deviations in predicted macroscopic physical responses \cite{ge20263d}.

In many imaging settings, lower-resolution 3D measurements are also readily available \cite{markelj2012review, deakin2026literature}. Carbon fiber-reinforced polymer (CFRP) composites provide a representative example. High-resolution techniques such as focused ion beam scanning electron microscopy (FIB-SEM) and synchrotron X-ray computed tomography (CT) can resolve individual fibers and their local arrangement, but their limited fields of view and low throughput restrict volumetric characterization to relatively small regions \cite{matsuda2026comparative, moschetti2026robot}. Laboratory micro-CT can cover substantially larger volumes at lower spatial resolution \cite{takahashi2022x}. Although these lower-resolution volumes cannot precisely resolve fiber-matrix interfaces, they retain specimen-specific information about spatial arrangement, continuity and larger-scale fiber architecture. This creates an opportunity to leverage fine-resolution 2D observations to improve the reconstruction of fine-scale 3D microstructures from lower-resolution volumetric data.

This setting can be formulated as a learning using privileged information (LUPI) problem, where fine-resolution 2D observations serve as privileged information available only during training, while lower-resolution 3D measurements are available during both training and inference \cite{vapnik2009new, hoffman2016learning}. Cross-modal knowledge distillation provides an effective framework for transferring structural knowledge between heterogeneous modalities, allowing a teacher to learn from fine-resolution 2D observations and guide a student that reconstructs fine-resolution 3D microstructures from lower-resolution 3D input data \cite{gou2021knowledge}. The main challenge is that the two modalities differ in imaging physics, spatial resolution and feature statistics, making reliable transfer between their representations difficult.

Cross-modal knowledge distillation can operate at different levels of supervision. Response-level distillation transfers knowledge by aligning predictions in a comparable output space, but its supervision is limited to the model outputs \cite{liu2023stereodistill, huo2024c, rakib2025g}. Feature-level distillation provides richer intermediate supervision, but direct feature alignment is sensitive to discrepancies between heterogeneous modalities \cite{chen2021learning, yang2024clip}. Relational knowledge distillation \cite{park2019relational, tung2019similarity, liu2019structured} offers another form of supervision by transferring relationships among learned representations. Recent cross-modal methods have explored sample-level relationships and spatial affinities \cite{wang2024multi, sun2026multi}. However, sample-level relations provide limited information about local spatial organization, while matching spatial affinities across modalities can still be affected by differences in their learned representations.
Feature decomposition provides another approach for addressing the modality gap, with existing methods using frequency decomposition or shared–private factorization to separate different components of learned features \cite{zhao2023spherical, zhang2024freekd, liu2026distilling, zhao2026pulse}. Such approaches rely on predefined decompositions to separate transferable and modality-specific information that may be intertwined within learned representations, which can lead to imperfect separation and the loss of useful structural information.

\begin{figure}[ht]
\centering
\includegraphics[width=\linewidth]{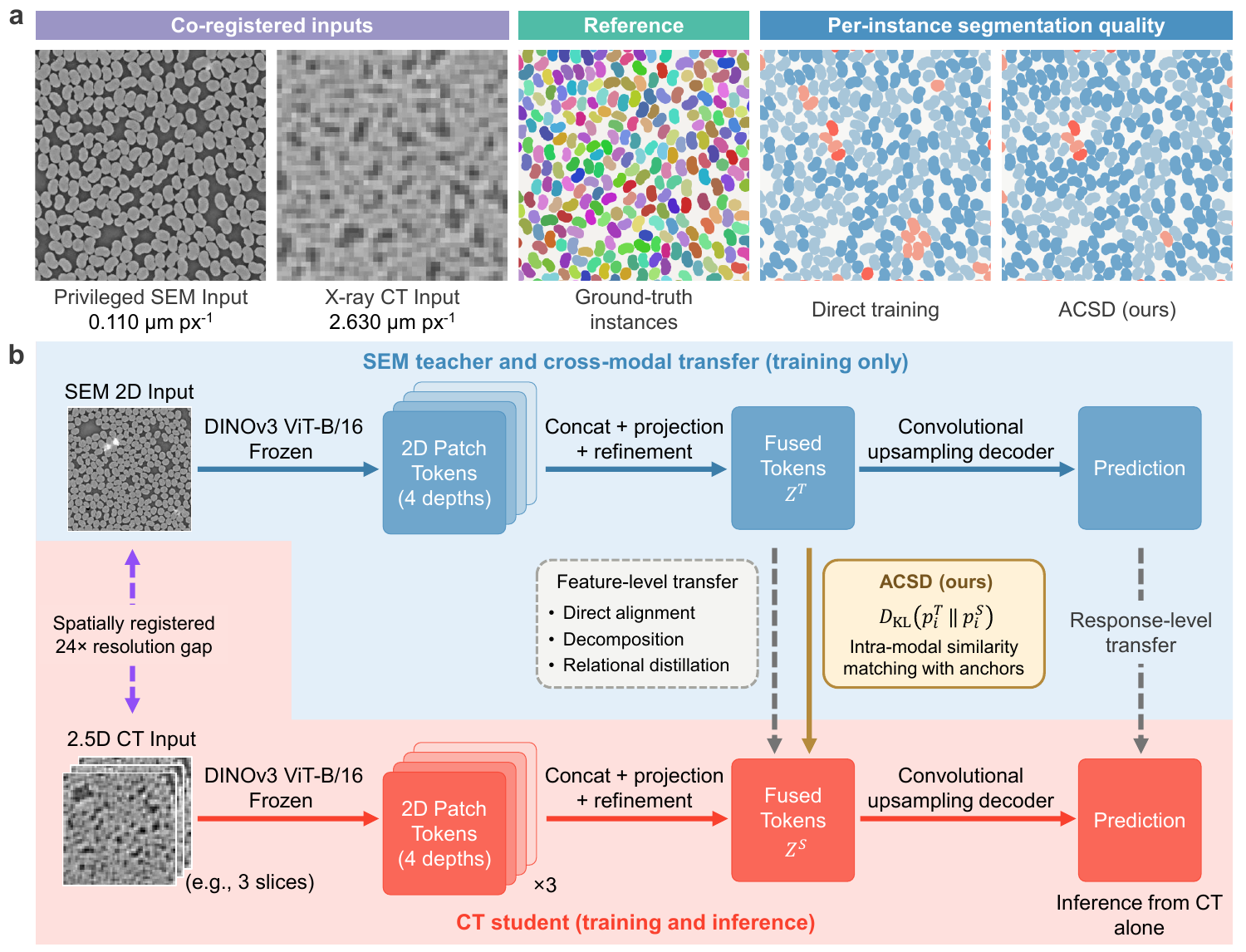}
\caption{\textbf{Privileged cross-modal relational distillation framework for microstructure reconstruction.} 
\textbf{a}, Representative example of imaging inputs, reference and reconstruction quality. Co-registered SEM and micro-CT images are shown with their corresponding pixel sizes, alongside the ground-truth fiber-instance reference and reconstruction quality visualizations for direct student training and ACSD. Each ground-truth fiber is colored according to the IoU of its matched prediction, with four IoU bands separated at 0.25, 0.50 and 0.75, progressing from coral for lower overlap to blue for higher overlap.
\textbf{b}, Overview of the PCRD framework. The SEM teacher and CT student extract and fuse features from four backbone depths for instance prediction. Gray paths summarize response-level and feature-level transfer, including direct feature alignment, feature decomposition and relational distillation. The gold path denotes ACSD-based transfer through anchor-conditioned intra-modal similarity distributions at registered spatial locations. The SEM teacher and cross-modal transfer are used only during training, while the CT student is retained for micro-CT-only inference.}
\label{fig:inputandmodel}
\end{figure}

We develop a Privileged Cross-modal Relational Distillation (PCRD) framework for specimen-specific 3D microstructure reconstruction using lower-resolution volumetric imaging, guided by fine-resolution 2D observations during training. As shown in Fig.~\ref{fig:inputandmodel}a, we apply PCRD to CFRP using co-registered fine-resolution SEM images and lower-resolution micro-CT measurements. As illustrated in Fig.~\ref{fig:inputandmodel}b, PCRD transfers spatial organization from a privileged SEM teacher to a CT student by capturing structural relationships within each modality-specific feature space, without directly matching heterogeneous feature vectors. We realize this idea through Anchored Cross-modal Similarity Distribution Distillation (ACSD), which represents teacher knowledge using similarity distributions between spatial anchors and the surrounding feature field. Our contributions are summarized as follows:

\begin{itemize}
    \item We formulate the specimen-specific 3D microstructure reconstruction as a LUPI problem, leveraging fine-resolution 2D observations during training to improve reconstruction from lower-resolution 3D measurements alone at inference.

    \item We introduce ACSD to transfer fine-scale spatial relationships from privileged 2D observations to the 3D student. By matching anchor-conditioned similarity distributions computed within each modality-specific feature space, ACSD enables knowledge transfer without direct feature alignment or predefined feature decomposition.

    \item We compare ACSD with prediction distillation, direct feature alignment, feature decomposition and cross-modal affinity distillation, and examine the effects of relational supervision, anchor selection and distillation settings through ablation studies. These experiments assess the contribution of similarity-distribution supervision and the design choices affecting knowledge transfer across modalities.

    \item We demonstrate PCRD on specimen-specific reconstruction of densely packed CFRP fibers. In 2D cross-sectional evaluation, PCRD improves mAP by 1.66 percentage points over direct micro-CT training. Independent 3D validation against high-resolution synchrotron X-ray CT data shows close agreement in fiber volume fraction, diameter and local spacing. The reconstruction matches 89.97\% of the reference fiber centerline length, compared with 75.08\% for the interpolated low-resolution volume.
\end{itemize}

\section{Results}\label{sec:2}
\subsection{Problem setting}

Unidirectional pultruded CFRP composites provide a demanding case study for 3D microstructure reconstruction, with our specimens having a fiber volume fraction of approximately 70\% and a mean fiber diameter of about 7~\(\mu\mathrm{m}\) \cite{mageira2026comprehensive}. As shown in Fig.~\ref{fig:inputandmodel}a, SEM clearly resolves individual fiber boundaries, while the corresponding micro-CT observation retains coarse spatial organization with limited boundary detail. The volumetric CT measurements also retain specimen-specific information about larger-scale fiber architecture and continuity, complementing the fine cross-sectional structure provided by SEM.

We formulate specimen-specific reconstruction as a LUPI problem. Each training example consists of a triplet \((x,x^{*},y)\), where \(x\) is the micro-CT observation, \(x^{*}\) is the spatially corresponding SEM observation available only during training, and \(y\) is the fine-resolution fiber-instance label map. Supervision is defined on registered 2D cross-sections, while the student uses adjacent micro-CT slices as a 2.5D input to incorporate volumetric context into the prediction. This design exploits the predominantly unidirectional fiber geometry, where the main reconstruction challenge lies in separating densely packed fibers in the transverse plane. The trained student is then applied section by section throughout the volume to reconstruct the 3D microstructure.

\subsection{Overview of the PCRD framework and ACSD}

PCRD consists of a privileged SEM teacher and a deployable micro-CT student operating on spatially registered observations with an approximately 24-fold difference in pixel size (Fig.~\ref{fig:inputandmodel}b). Before feature extraction, the micro-CT input is interpolated onto the registered SEM grid. Both branches use frozen DINOv3 ViT-B/16 \cite{siméoni2025dinov3} encoders to extract 2D patch-token representations from four backbone depths. The multiscale features are concatenated and projected through a \(1\times1\) layer to form fused teacher and student representations. Convolutional upsampling decoders then predict fiber foreground probabilities and flow fields \cite{stringer2021cellpose} that guide the separation of individual fiber instances. During training, ACSD transfers structural knowledge from the SEM teacher to the micro-CT student at the fused representation level. During inference, only the micro-CT student is retained. Architectural and optimization details are provided in Methods~\ref{method:PCRD}.

\begin{figure}[ht]
\centering
\includegraphics[width=\linewidth]{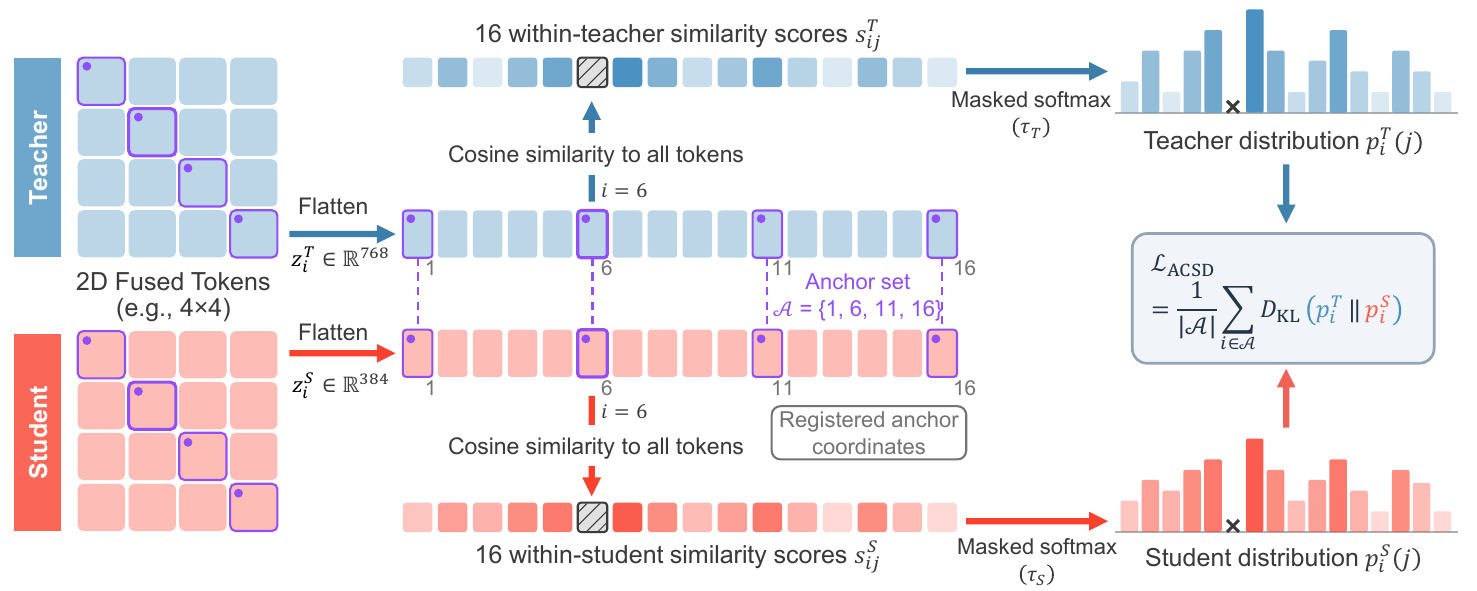}
\caption{\textbf{Anchored cross-modal similarity distribution distillation.} A \(4\times4\) token grid is shown schematically for clarity. The fused teacher and student feature grids are flattened into token sequences, with spatially corresponding anchors outlined in purple and marked by purple dots. For each anchor, cosine similarities with the remaining tokens within the same feature space are converted by masked softmax into teacher and student similarity distributions. ACSD minimizes the KL divergence between corresponding distributions across the shared anchor set, transferring spatial organization from the privileged SEM teacher to the CT student.}
\label{fig:ACSD}
\end{figure}

ACSD transfers knowledge through the spatial organization encoded within each modality-specific representation (Fig.~\ref{fig:ACSD}). The fused teacher and student feature fields are first flattened into spatial tokens, and a shared set of spatially corresponding locations is selected as anchors. The limited spatial resolution of micro-CT can cause neighboring feature locations to encode similar structural information, making dense anchor supervision potentially redundant. ACSD therefore selects a subset of anchors to reduce redundant constraints and computational cost, while each anchor still establishes relationships with every spatial token in the feature field. Within each modality, cosine similarities between each anchor and all spatial tokens are converted into an anchor-conditioned distribution using masked softmax, which excludes the anchor's self-similarity. Separate teacher and student temperatures control the concentration of their respective similarity distributions, allowing the relational target and student response to be adjusted independently. ACSD then matches the corresponding teacher and student distributions, transferring the spatial relationships encoded relative to each anchor without directly aligning the underlying SEM and micro-CT feature vectors. The complete formulation of the similarity distributions and distillation objective is provided in Methods~\ref{method:ACSD}.

\subsection{Relational transfer improves fiber-instance reconstruction}

We evaluated the knowledge-transfer strategies on 3,200 spatially registered micro-CT--SEM pairs, with an average of 297 annotated fiber instances per sample. We assessed 2D cross-sectional reconstruction performance using mean average precision (mAP), F1 scores at IoU thresholds of 0.50 and 0.75, and Binary IoU. mAP summarizes the precision–recall performance of fiber-instance predictions across multiple matching thresholds. F1 scores capture the balance between correctly identified, missed and falsely detected fibers at fixed matching criteria, with F1@0.5 assessing instance identification under a more permissive overlap threshold and F1@0.75 requiring closer agreement with reference instances. Binary IoU measures agreement of the overall fiber phase, independent of whether individual fibers are correctly separated. Full evaluation details are provided in Methods~\ref{method:training}.

To examine how different forms of privileged supervision affect fiber-instance reconstruction, we adapted five existing distillation strategies for comparison with ACSD and direct student training, which uses only ground-truth supervision. All methods share the same student backbone, decoder and training protocol, with feature-based distillation applied at the fused representation level. Prediction distillation aligns the teacher and student foreground and flow predictions, while direct feature alignment constrains corresponding teacher and student fused features to match. Frequency decomposition extracts and aligns low-frequency feature components, which are expected to capture coarse structural patterns shared across the resolution gap \cite{zhang2024freekd,liu2026distilling}. Shared-private decomposition factorizes the fused features into modality-shared and modality-specific components, transferring knowledge through alignment of the shared representations \cite{bousmalis2016domain,van2023aspnet}. Cross-affinity distillation, adapted from a recent cross-modal distillation method for dense prediction \cite{sun2026multi}, matches cross-modal teacher–student affinities with student intra-modal affinities to transfer spatial relationships, but still involves direct similarity computation between heterogeneous feature vectors. Implementation details of the adapted baselines are provided in the Supplementary Information.

ACSD improved fiber-instance reconstruction from micro-CT inputs across all four evaluation metrics (Table~\ref{tab:main_results}). Relative to direct training, ACSD increased mAP by 1.66 percentage points and F1@0.75 by 2.65 points, while F1@0.5 improved by 0.61 points. Binary IoU increased by 0.26 percentage points, while the larger gains in instance-level metrics suggest that ACSD improves individual fiber reconstruction beyond overall fiber-phase segmentation. The example in Fig.~\ref{fig:inputandmodel}a further illustrates this improvement, with ACSD achieving higher per-instance overlap than direct student training.

\begin{table}[ht]
\centering
\setlength{\tabcolsep}{2.5pt}
\footnotesize
\caption{Comparison of cross-modal knowledge distillation strategies for CFRP fiber-instance reconstruction. Results are reported as mean $\pm$ standard deviation over five runs with different random seeds.}
\label{tab:main_results}
\begin{tabular}{lcccc}
\toprule
Method
& mAP $\uparrow$
& F1@0.5 $\uparrow$
& F1@0.75 $\uparrow$
& Binary IoU $\uparrow$ \\
\midrule

Direct training
& $0.4253 \pm 0.0014$
& $0.9184 \pm 0.0017$
& $0.4973 \pm 0.0011$
& $0.8208 \pm 0.0007$ \\

\midrule

Prediction distillation
& $0.4254 \pm 0.0020$
& $0.9190 \pm 0.0008$
& $0.4989 \pm 0.0039$
& $0.8209 \pm 0.0009$ \\

Feature alignment
& $0.4288 \pm 0.0019$
& $0.9192 \pm 0.0012$
& $0.5036 \pm 0.0029$
& $0.8210 \pm 0.0006$ \\

Frequency decomposition
& $0.4259 \pm 0.0021$
& $0.9181 \pm 0.0012$
& $0.4982 \pm 0.0014$
& $0.8201 \pm 0.0009$ \\

Shared-private decomposition
& $0.4317 \pm 0.0013$
& $0.9204 \pm 0.0007$
& $0.5064 \pm 0.0032$
& $0.8221 \pm 0.0004$ \\

Cross-affinity distillation
& $0.4239 \pm 0.0003$
& $0.9186 \pm 0.0011$
& $0.4957 \pm 0.0005$
& $0.8213 \pm 0.0005$ \\

\midrule

ACSD (ours)
& $\mathbf{0.4419 \pm 0.0007}$
& $\mathbf{0.9245 \pm 0.0008}$
& $\mathbf{0.5238 \pm 0.0020}$
& $\mathbf{0.8234 \pm 0.0006}$ \\

\bottomrule
\end{tabular}
\end{table}

The comparison with alternative distillation strategies highlights the importance of how privileged knowledge is represented and transferred. Prediction distillation produced only a marginal improvement over direct training, increasing mAP by 0.01 percentage points. Direct feature alignment yielded a somewhat larger, but still limited gain of 0.35 points. Frequency decomposition improved mAP by only 0.06 points, although low-frequency features may capture coarse structural patterns shared between SEM and micro-CT observations. Shared-private decomposition performed best among the feature-based baselines, improving mAP by 0.64 points. Cross-affinity distillation performed slightly below direct training and was also outperformed by direct feature alignment, suggesting that its formulation of relational knowledge may be more sensitive to the modality gap than direct feature matching. ACSD outperformed the strongest decomposition baseline by 1.02 mAP points and 1.74 F1@0.75 points. These results support the effectiveness of transferring spatial relationships through intra-modal similarity distributions for fine-scale fiber-instance reconstruction.

\subsection{Analysis of relational transfer}

\begin{figure}[ht]
\centering
\includegraphics[width=\linewidth]{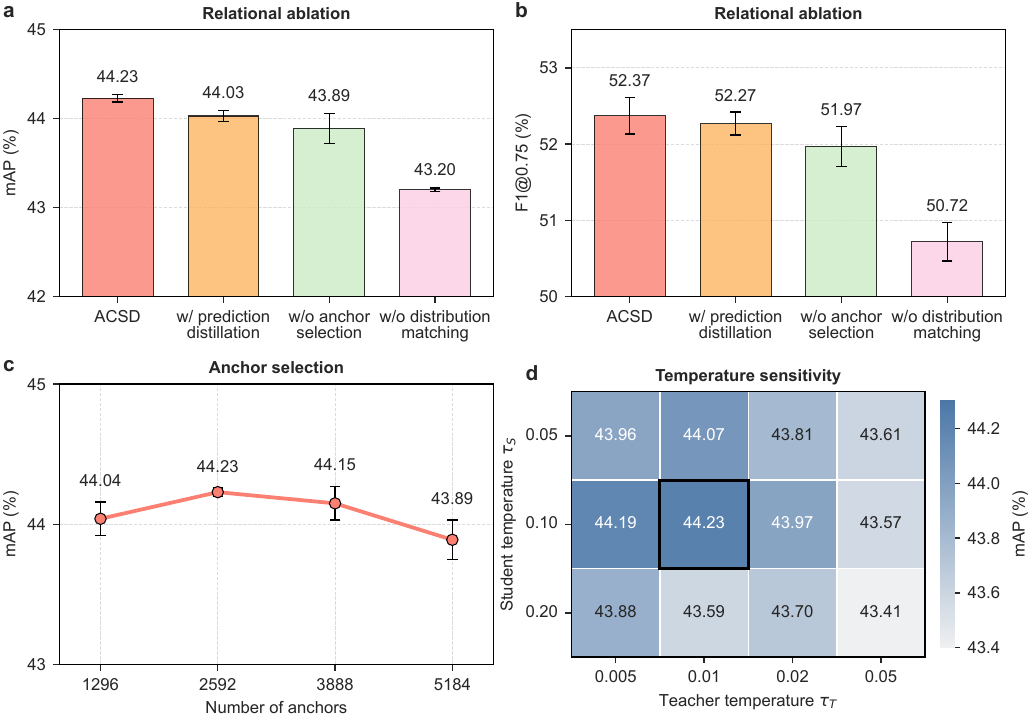}
\caption{\textbf{Analysis of relational transfer in ACSD.} \textbf{a,b}, Effects of similarity-distribution matching, prediction distillation and anchor selection on mAP and F1@0.75. \textbf{c}, Effect of the number of spatial anchors on mAP. \textbf{d}, Sensitivity of mAP to the teacher temperature \(\tau_T\) and student temperature \(\tau_S\). Results are reported as means over three runs with different random seeds, with standard deviation error bars shown in a, b and c.}
\label{fig:relational_analysis}
\end{figure}

We first examined whether the similarity distribution itself was responsible for the improvement in relational transfer. In ACSD, selected spatial locations serve as anchors for measuring their similarity to the entire feature field. The ACSD w/o distribution matching variant retains anchor-based supervision but replaces the full similarity distribution with hard relations, where the teacher selects highly similar and dissimilar patch locations for each anchor and supervises the student through a contrastive objective. As shown in Fig.~\ref{fig:relational_analysis}a,b, removing distribution matching reduced mAP by 1.03 percentage points and F1@0.75 by 1.65 points. These results support the importance of preserving graded similarity information, as reducing teacher relations to positive and negative sets discards information about the relative organization of spatial locations.

We also tested whether output-level teacher supervision provided additional information beyond the relational objective. Adding prediction distillation to ACSD reduced mAP by 0.20 percentage points and F1@0.75 by 0.10 points (Fig.~\ref{fig:relational_analysis}a,b). These results suggest that prediction-level supervision provides little additional benefit when combined with ACSD in this setting.

We next examined the effect of anchor selection on relational transfer (Fig.~\ref{fig:relational_analysis}a–c). Removing anchor selection, which uses all 5,184 spatial locations as anchors, reduced mAP by 0.34 percentage points and F1@0.75 by 0.40 points relative to ACSD. To investigate how anchor density affects reconstruction, we also evaluated smaller anchor sets. Increasing the number of anchors from 1,296 to 2,592 improved mean mAP by 0.19 percentage points, while increasing it to 3,888 reduced mAP by 0.08 points relative to the 2,592-anchor setting. Together with the decline observed when all 5,184 locations were used as anchors, these results show that using more anchors does not necessarily improve reconstruction accuracy.

Finally, we examined the role of separate teacher and student temperatures in relational transfer (Fig.~\ref{fig:relational_analysis}d). The teacher temperature controls the concentration of the teacher similarity distribution. A lower temperature produces a sharper distribution, assigning greater probability to spatial locations that are more similar to each anchor. The student temperature independently controls the smoothness of the student distribution, allowing greater flexibility in matching teacher relations across the modality gap. To examine their effects, we varied \(\tau_T\) over \(\{0.005,0.01,0.02,0.05\}\) and \(\tau_S\) over \(\{0.05,0.1,0.2\}\). Relatively low teacher temperatures combined with a moderately higher student temperature generally yielded better reconstruction performance. Increasing \(\tau_T\) to 0.05 reduced mAP in most settings, while increasing \(\tau_S\) to 0.2 consistently reduced performance across all teacher temperatures. These results support the use of separate temperatures to maintain a concentrated teacher relational target while allowing the student to accommodate cross-modal representation differences without excessive smoothing.

\subsection{3D microstructure reconstruction and validation}

\begin{figure}[ht]
\centering
\includegraphics[width=\linewidth]{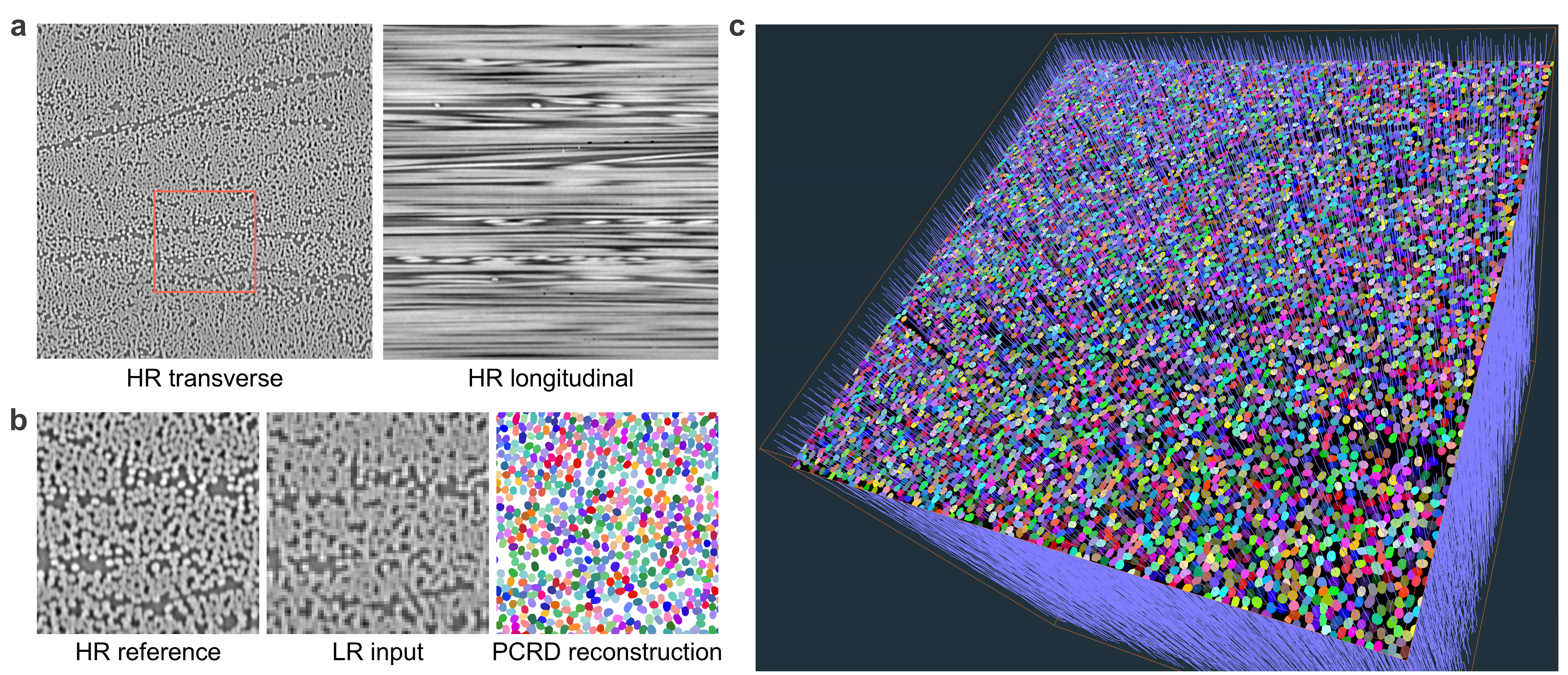}
\caption{\textbf{3D reconstruction and validation against synchrotron X-ray CT data.} \textbf{a}, Transverse and longitudinal views of the HR reference, with red boxes indicating the region selected for 3D validation. \textbf{b}, Enlarged transverse region showing the HR reference, LR input and PCRD reconstruction. Individual fibers that are poorly resolved in the downsampled input are recovered as distinct instances after reconstruction. \textbf{c}, 3D visualization of the PCRD reconstruction and the fiber centerlines extracted by the subsequent tracking analysis, which are used to quantify fiber trajectory fidelity through the volume.}
\label{fig:3d_validation}
\end{figure}

To assess whether the reconstructed cross-sections reproduce the underlying 3D architecture, we performed an independent volumetric validation using high-resolution synchrotron X-ray CT data as the reference (HR reference). A low-resolution volume (LR input) of \(248^3\) voxels, covering approximately \(0.652~\mathrm{mm}\) on each side, was generated from the HR reference at the same voxel size as the micro-CT training data, following established downsampling procedures \cite{cardoso2015scale,diaz2017downsampling,sjolin2017compression}. The selected validation region is highlighted in the transverse and longitudinal views of the HR reference in Fig.~\ref{fig:3d_validation}a. The trained micro-CT student was applied to the LR input to generate fiber instance masks, which were assembled into a 3D volume referred to as the PCRD reconstruction. We compared the PCRD reconstruction and an interpolated version of the LR input (interpolated LR) against the HR reference using structural statistics. Fiber volume fraction measures the proportion of the volume occupied by fibers, while equivalent fiber diameter characterizes individual fiber size through the diameter of a circle with the same cross-sectional area. Nearest-neighbor distance describes local fiber packing by measuring the distance between each fiber and its closest neighboring fiber.

The PCRD reconstruction closely reproduced the structural statistics of the HR reference (Table~\ref{tab:microstructure_validation}). The reconstructed fiber volume fraction was 71.85\%, compared with 71.89\% in the HR reference and 74.52\% in the interpolated LR. The median equivalent fiber diameter was 7.33~\(\mu\mathrm{m}\) for the PCRD reconstruction and 7.35~\(\mu\mathrm{m}\) for the HR reference. Median nearest-neighbor distances were also similar, at 6.83 and 6.61~\(\mu\mathrm{m}\), respectively. 
Reliable estimates of fiber diameter and nearest-neighbor distance could not be obtained from the interpolated LR because individual fibers remained insufficiently resolved. Figure~\ref{fig:3d_validation}b further illustrates the improvement in fiber separation, with distinct fiber instances recovered in the PCRD reconstruction.

\begin{table}[ht]
\centering
\setlength{\tabcolsep}{2pt}
\caption{Structural properties and fiber tracking fidelity relative to the high-resolution synchrotron X-ray CT reference.}
\label{tab:microstructure_validation}
\begin{tabular}{lccc}
\toprule
Metric & HR reference & Interpolated LR & PCRD reconstruction \\
\midrule

Fiber volume fraction (\%)
& 71.89
& 74.52
& 71.85 \\

Equivalent fiber diameter ($\mu$m)
& 7.35 [6.93, 7.74]
& --
& 7.33 [7.08, 7.58] \\

Nearest-neighbor distance ($\mu$m)
& 6.61 [6.20, 7.01]
& --
& 6.83 [6.48, 7.20] \\

\midrule

Reference centerline coverage (\%) $\uparrow$
& Ref.
& 75.08
& 89.97 \\

Centerline position deviation ($\mu$m) $\downarrow$
& Ref.
& 0.77 [0.44, 1.32]
& 0.98 [0.63, 1.52] \\

Orientation deviation ($^\circ$) $\downarrow$
& Ref.
& 0.78 [0.00, 2.49]
& 1.12 [0.00, 2.49] \\

\bottomrule
\end{tabular}

\vspace{2pt}
\footnotetext{
Equivalent fiber diameter and nearest-neighbor distance are reported as median [IQR].
Tracking deviations are computed relative to the HR reference and reported as length-weighted median [IQR].
A dash indicates that reliable instance-level measurements could not be obtained from the interpolated LR volume.}
\end{table}

To further assess fiber trajectory recovery and geometric fidelity across the reconstructed volume, we extracted 3D fiber centerlines and compared them with those of the HR reference. The PCRD reconstruction and its extracted fiber centerlines are shown in Fig.~\ref{fig:3d_validation}c. Reference centerline coverage measures the proportion of the total HR reference centerline length that can be matched to reconstructed fiber trajectories. Position and orientation deviations quantify the spatial displacement and local directional differences between matched centerlines, respectively. The PCRD reconstruction matched 89.97\% of the reference centerline length, compared with 75.08\% for the interpolated LR, an increase of 14.89 percentage points. The higher coverage indicates that the PCRD reconstruction recovers a greater proportion of the reference fiber trajectories across the validation volume. For the matched centerline segments, the PCRD reconstruction yielded a length-weighted median position deviation of 0.98~\(\mu\mathrm{m}\) and an orientation deviation of 1.12\(^{\circ}\), compared with 0.77~\(\mu\mathrm{m}\) and 0.78\(^{\circ}\) for the interpolated LR.

Together, the structural and tracking analyses show that PCRD recovers a coherent 3D fiber architecture with close agreement to the HR reference over the validation volume. The reconstructed microstructure can support subsequent geometric characterization and physics-based modelling \cite{bikos2025simulating} using low-resolution volumetric data.

\section{Discussion}\label{sec:3}

This study shows that privileged high-resolution 2D observations can improve specimen-specific reconstruction from a lower-resolution volumetric modality when supervision is transferred through the spatial organization of learned representations. In the CFRP system studied here, SEM resolves fine-scale fiber–matrix interfaces that are poorly resolved in micro-CT images, while micro-CT retains information about the 3D arrangement of the specimen. PCRD uses these fine-resolution 2D observations only during training, with inference relying solely on micro-CT. Across the tested transfer strategies, ACSD produced the largest improvement in fiber-instance reconstruction. These findings support the use of within-modality spatial relationships for knowledge transfer between imaging modalities with different image formation processes.

The comparison with other distillation strategies helps explain this result. Teacher predictions provide task-specific supervision, but contain less information than the intermediate SEM representation from which they are generated. Direct feature alignment assumes that corresponding teacher and student features are comparable, an assumption that may not hold across different imaging modalities. Frequency and shared--private decomposition seek to address these differences by imposing additional structure on the representations before alignment, but their improvements remained below those achieved by ACSD. Cross-affinity distillation also transfers spatial relationships, but its use of cross-modal feature similarities may retain sensitivity to differences between the teacher and student representation spaces. ACSD instead constructs similarity distributions independently within each modality and matches the resulting spatial relationships. This allows the teacher and student to retain different feature representations without specifying which feature components should be shared across modalities.

The relational analyses provide further insight into the design of ACSD. Replacing the similarity distribution with hard positive and negative relations reduced reconstruction accuracy, indicating that graded similarity information contributes to relational transfer. Using more anchors did not consistently improve reconstruction accuracy. A moderate anchor density may balance coverage of spatial structures against redundant relational constraints, particularly when neighboring micro-CT features encode similar information due to limited spatial resolution. The temperature analysis also showed that the teacher and student distributions benefit from different levels of concentration. A sharper teacher distribution emphasizes the most similar spatial locations, while a moderately smoother student distribution may help accommodate differences between the two representation spaces.

The independent synchrotron X-ray CT validation shows that the benefits of PCRD extend beyond individual cross-sections. The reconstructed volume closely reproduced the fiber volume fraction, equivalent fiber diameter and local spacing of the HR reference. PCRD also recovered a greater proportion of the reference centerline length than the interpolated LR, although its position and orientation deviations were slightly higher on the matched segments. The low-resolution volumetric input provides specimen-specific spatial information, while privileged supervision helps recover individual fiber structures that are difficult to resolve from that input alone. The ability to reconstruct fine-scale fiber geometry from lower-resolution volumetric data could support 3D image-based modelling over larger specimen regions, with potential applications in manufacturing process optimization and material property prediction.

The formulation of ACSD is not specific to CFRP or the SEM-CT imaging pair and may be applicable to other imaging systems with spatially corresponding observations. More generally, ACSD requires spatial correspondence between the two modalities and informative relationships among locations within each modality. Two limitations remain in the present study. First, the reconstruction strategy improves spatial resolution primarily in the transverse plane, while resolution along the fiber direction remains constrained by the micro-CT sampling. Its ability to recover isotropically high-resolution 3D microstructures has therefore not been established. A possible extension is to reconstruct along multiple directions and combine the resulting predictions, with direction-specific training where required by material anisotropy. Second, ACSD constructs spatial relationships within individual observations. The densely packed fiber system studied here provides many similar local structures, whereas sparse or highly heterogeneous microstructures may offer fewer informative relationships for within-sample distillation. The effectiveness of ACSD in these settings remains to be evaluated.

\section{Methods}\label{sec:4}

\subsection{Data preparation and cross-modal registration}

Two unidirectional pultruded CFRP specimens with a vinyl ester matrix were prepared for paired SEM and micro-CT imaging. Each specimen measured approximately \(5\times5\times10\) mm, with one end face polished for cross-sectional imaging. Volumetric micro-CT data were acquired using a Thermo Scientific HeliScan microCT system at an isotropic voxel size of \(2.630~\mu\mathrm{m}\). Cross-sectional SEM images were acquired using a Thermo Scientific Phenom XL G2 Desktop SEM with a backscattered electron detector at a pixel size of \(0.098~\mu\mathrm{m}\).

To establish spatial correspondence between the two modalities, a representative micro-CT end-face image was constructed from the reconstructed volume to account for variations in the polished surface depth. Registration was performed in two stages: an initial affine alignment of the full SEM and micro-CT end-face images, followed by local rigid registration of corresponding image regions. The registered SEM images were resampled to approximately \(0.110~\mu\mathrm{m\,px^{-1}}\) and cropped into \(1152\times1152\)-pixel patches, each corresponding to a \(48\times48\)-pixel micro-CT region on its native sampling grid. Further details of specimen preparation and cross-modal registration are provided in the Supplementary Information.

Fiber-instance annotations were generated from the registered SEM images using Cellpose-SAM, followed by label filtering to remove detections inconsistent with the expected fiber morphology. All image samples were manually inspected and corrected at the instance level. Each retained fiber cross-section was assigned an independent instance label. In total, 3,200 registered SEM–micro-CT pairs were obtained from the two specimens, with each pair representing a distinct cross-sectional region. The supervised target corresponds to the registered specimen end face, and the 2.5D student input therefore uses consecutive micro-CT slices extending into the specimen. The same one-sided input arrangement was used for interior sections during volumetric inference to maintain consistency with training.

For independent 3D validation, we used a previously reported synchrotron X-ray CT dataset of the same material system \cite{mageira2026comprehensive}. The LR input was generated from the HR reference by blurring and downsampling to the experimental micro-CT voxel size of \(2.630~\mu\mathrm{m}\), following established procedures \cite{cardoso2015scale,diaz2017downsampling,sjolin2017compression}. A \(248^3\)-voxel region, covering approximately \(0.652~\mathrm{mm}\) on each side, was selected for reconstruction. The same LR input was interpolated onto the high-resolution grid to construct the interpolated LR baseline. Further details of downsampling and intensity calibration are provided in the Supplementary Information.

\subsection{Privileged cross-modal relational distillation framework}\label{method:PCRD}

PCRD consists of an SEM teacher and a micro-CT student operating on spatially registered observations (Fig.~\ref{fig:inputandmodel}b). The teacher receives the SEM observation \(x^{*}\) and is used only during training, while the student receives the corresponding micro-CT observation \(x\) and is retained for inference. Both branches use frozen DINOv3 ViT-B/16 \cite{siméoni2025dinov3} encoders with the same pretrained backbone architecture. Before feature extraction, the micro-CT input is interpolated onto the registered SEM grid so that corresponding in-plane physical locations are mapped to aligned patch lattices in the teacher and student encoders.
We extract spatial token representations from transformer blocks 3, 6, 9 and 12, corresponding to 25\%, 50\%, 75\% and 100\% of the DINOv3 backbone depth, respectively,
\[
\{F_{\ell}^{T}\}_{\ell=1}^{4}=\phi_T(x^{*}),
\quad
\{F_{\ell}^{S}\}_{\ell=1}^{4}=\phi_S(x),
\]
and concatenate the four feature fields along the channel dimension to form fused spatial representations,
\[
Z^{T}
=
g_T\!\left(
\operatorname{Concat}_{\ell=1}^{4}F_{\ell}^{T}
\right),
\quad
Z^{S}
=
g_S\!\left(
\operatorname{Concat}_{\ell=1}^{4}F_{\ell}^{S}
\right).
\]
Here, \(g_T\) and \(g_S\) consist of a \(1\times1\) convolution, GroupNorm and GELU activation followed by two residual convolutional blocks. For the 2.5D student input, features extracted from three consecutive micro-CT slices are concatenated at each backbone depth and projected through a depth-specific \(1\times1\) convolution before multiscale fusion. The fused teacher and student representations contain 768 and 384 channels, respectively. The lower-dimensional student representation was chosen to encourage a more compact task representation for deployment while retaining sufficient capacity for reconstruction from lower-resolution micro-CT observations. The student feature dimension and number of input slices were selected through architectural ablations reported in the Supplementary Information.

Each fused representation is decoded by a convolutional upsampling pathway that progressively restores spatial resolution and predicts a foreground probability map and a two-component instance flow field. Following the dense flow representation used in Cellpose \cite{stringer2021cellpose}, each foreground pixel is assigned a vector directed toward the interior of its corresponding fiber instance. The foreground output is supervised using binary cross-entropy, while the flow field is supervised using an \(L_2\) loss. We define the reconstruction objective as
\[
\mathcal{L}_{\mathrm{rec}}
=
\lambda_{\mathrm{fg}}\mathcal{L}_{\mathrm{fg}}
+
\lambda_{\mathrm{flow}}\mathcal{L}_{\mathrm{flow}}.
\]
The SEM teacher is first optimized using \(\mathcal{L}_{\mathrm{rec}}\) and is then fixed during student training. The student retains the same foreground and flow reconstruction objective, with an additional cross-modal relational term introduced at the fused representation level. Specifically, ACSD is applied to \(Z^{T}\) and \(Z^{S}\) before decoding to transfer the spatial organization learned by the teacher. At inference, only the trained student is retained for section-wise reconstruction from micro-CT data.

\subsection{Anchored cross-modal similarity distribution distillation}\label{method:ACSD}

ACSD operates on the fused teacher and student representations \(Z^{T}\) and \(Z^{S}\) and transfers spatial organization through intra-modal similarity distributions. Because the limited spatial resolution of micro-CT can cause neighboring feature locations to encode similar structural information, using every location as an anchor may introduce redundant relational constraints. We therefore select a subset of spatial anchors to reduce redundancy and computational cost, while retaining comparisons between each anchor and the complete feature field. The teacher and student use the same anchor locations through spatial registration, with similarities computed independently within each representation space.

We first flatten each feature field over its spatial dimensions and \(L_2\)-normalize the feature vector at each patch location, yielding \(z_i^{T}\) and \(z_i^{S}\) for spatial location \(i\) as shown in Fig.~\ref{fig:ACSD}. We then select a set of spatial anchors \(\mathcal A\) at uniformly spaced positions along the flattened token sequence. For each anchor \(i\in\mathcal A\), we compute its cosine similarity to every patch location \(j\),
\[
s_{ij}^{T}=\left\langle z_i^{T},z_j^{T}\right\rangle,
\quad
s_{ij}^{S}=\left\langle z_i^{S},z_j^{S}\right\rangle .
\]
The resulting similarities are converted into anchor-conditioned distributions,
\[
p_i^{T}(j)
=
\frac{\exp\!\left(s_{ij}^{T}/\tau_T\right)}
{\sum_{k\neq i}\exp\!\left(s_{ik}^{T}/\tau_T\right)},
\quad
p_i^{S}(j)
=
\frac{\exp\!\left(s_{ij}^{S}/\tau_S\right)}
{\sum_{k\neq i}\exp\!\left(s_{ik}^{S}/\tau_S\right)},
\qquad j\neq i,
\]

where the anchor self-similarity is masked before normalization and \(\tau_T\) and \(\tau_S\) independently control the concentration of the teacher and student similarity distributions. A lower teacher temperature sharpens the relational target and emphasizes the most structurally similar locations, whereas a higher student temperature yields a smoother response that is less sensitive to cross-modal feature discrepancies. The two temperatures are therefore decoupled to accommodate the different roles of teacher supervision and student adaptation.

The student is trained to reproduce the teacher relational distributions by minimizing
\[
\mathcal{L}_{\mathrm{ACSD}}
=
\frac{1}{|\mathcal A|}
\sum_{i\in\mathcal A}
D_{\mathrm{KL}}
\!\left(
p_i^{T}\,\|\,p_i^{S}
\right),
\]
where \(D_{\mathrm{KL}}(\cdot\|\cdot)\) denotes the Kullback--Leibler divergence and the teacher similarity distribution serves as the fixed target. ACSD is added to the student reconstruction objective,
\[
\mathcal{L}_{S}
=
\mathcal{L}_{\mathrm{rec}}^{S}
+
\lambda_{\mathrm{ACSD}}\mathcal{L}_{\mathrm{ACSD}},
\]
where \(\lambda_{\mathrm{ACSD}}\) controls the contribution of relational supervision. During student optimization, all teacher parameters remain fixed.

\subsection{Experimental setup and evaluation}\label{method:training}

The 3,200 registered SEM-micro-CT pairs were randomly split into training, validation and test sets in an 8:1:1 ratio, stratified by specimen to maintain equal representation of the two specimens in each subset. The split was fixed across all experiments so that all knowledge-transfer strategies were evaluated on the same samples. We compared ACSD with direct student training and five distillation strategies: prediction distillation, direct feature alignment, frequency decomposition, shared-private decomposition and cross-affinity distillation. All methods used the same student backbone, decoder, reconstruction targets and optimization protocol, with method-specific distillation components and objectives described in the Supplementary Information.

The SEM teacher was trained first and then fixed during student optimization. Teacher predictions and fused representations required by the different distillation strategies were precomputed and cached for the corresponding training samples. Teacher and student models were optimized with AdamW for 16 epochs using a base learning rate of \(7\times10^{-4}\) and a weight decay of \(10^{-4}\). The learning rate was linearly warmed up for the first three epochs and then decayed using a cosine schedule to \(0.1\) times the base learning rate. Training used bfloat16 mixed precision. The foreground and flow loss weights were set to \(\lambda_{\mathrm{fg}}=1\) and \(\lambda_{\mathrm{flow}}=0.4\), respectively, and ACSD used \(\lambda_{\mathrm{ACSD}}=1\). Unless otherwise stated, ACSD used 2,592 uniformly spaced anchors over the flattened token sequence, with teacher and student temperatures of \(\tau_T=0.01\) and \(\tau_S=0.1\), respectively. All methods were evaluated using a common set of predefined random seeds, with data-loader initialization controlled for reproducibility. Main comparisons were repeated across five independent training runs, while relational ablations and sensitivity analyses used three runs.

Instance masks were recovered from the predicted foreground logits and flow fields using flow-based grouping. Pixels above a foreground-logit threshold of 0 were propagated along the predicted flow field for 200 iterations and grouped according to their converged locations. Reconstructed instances smaller than 1,000 pixels were removed. Ground-truth instances intersecting the image boundary were excluded from instance-level evaluation because only part of the corresponding fiber was observed within the cropped image tile. Predicted instances whose area overlapped an excluded boundary instance by at least 50\% were also removed so that correct reconstruction of a cropped fiber was not counted as a false positive.

Reconstruction accuracy was evaluated over IoU thresholds from 0.50 to 0.95 in increments of 0.05. At each threshold, predicted and ground-truth instances were matched one-to-one according to IoU. We report F1 scores at IoU thresholds of 0.50 and 0.75, together with COCO-style mean average precision (mAP) computed across the full threshold range using the mean predicted foreground probability within each reconstructed instance as its confidence score. Binary IoU was computed independently of instance identity to quantify agreement of the total reconstructed fiber phase. Hyperparameters, including the number of anchors and the teacher and student temperatures, were selected based on validation-set performance. For each training run, the checkpoint with the highest validation mAP was used for evaluation on the held-out test set. The test set was not used for hyperparameter or checkpoint selection.

\subsection{Volumetric inference and structural analysis}\label{method:3d}

The trained student was applied section by section to the LR input using the same 2.5D input construction as during training. Each target section was reconstructed from overlapping tiles, with predicted flow fields and foreground logits blended in the overlap regions using cosine-weighted averaging. The stitched predictions were converted into fiber-instance masks and assembled into a 3D volume. The resulting \(5952\times5952\times248\) binary mask stack was downsampled in the transverse plane using pooling and thresholding, followed by resampling along the fiber direction to obtain a \(1488^3\) isotropic binary volume with a voxel size of approximately \(0.438~\mu\mathrm{m}\) for fiber tracking.

Fiber volume fraction was calculated as the proportion of the selected region of interest (ROI) occupied by fibers. For the HR reference and interpolated LR, the fiber phase was identified using entropy-based thresholding \cite{kapur1985new}, applied at their respective analysis resolutions. For the PCRD reconstruction, fiber volume fraction was calculated directly from the predicted binary masks. To obtain individual fiber measurements from the HR reference, fiber instances were segmented using Cellpose-SAM and filtered according to the expected fiber morphology. Equivalent fiber diameter was calculated from the cross-sectional area of each instance, and nearest-neighbor distance was defined as the Euclidean distance between each fiber centroid and its closest neighboring fiber centroid. The same measurements were obtained directly from the predicted instances in the PCRD reconstruction.

Fiber centerlines were extracted from the three volumes using cylindrical template matching followed by bidirectional tracing guided by local correlation and orientation consistency. The extracted centerlines were evaluated against those of the HR reference. Centerlines were interpolated onto common axial planes and matched using mutual nearest neighbors within \(3~\mu\mathrm{m}\). Reference centerline coverage was defined as the proportion of the total reference centerline length for which both endpoints of a sampled segment were successfully matched. Position deviation was calculated from the transverse distance between matched centerline locations, while orientation deviation was calculated from the angle between their local tangent directions. Position and orientation deviations were evaluated over the matched portions of each reconstruction independently and reported as reference-length-weighted median [IQR]. Further details of structural characterization, template matching and centerline evaluation are provided in the Supplementary Information.

\backmatter


\bmhead{Acknowledgements}
The authors acknowledge funding from Horizon Europe through the MSCA Doctoral Network RELIANCE, grant no. 101073040. 




\noindent

\bibliography{sn-bibliography}

\newpage
\section*{Supplementary information}

\setcounter{figure}{0}
\setcounter{table}{0}
\renewcommand{\thefigure}{S\arabic{figure}}
\renewcommand{\thetable}{S\arabic{table}}

\renewcommand{\theHfigure}{S\arabic{figure}}
\renewcommand{\theHtable}{S\arabic{table}}

\subsection*{Note 1: Hardware and computational environment}

All deep-learning experiments were conducted on a workstation equipped with a single NVIDIA RTX PRO 6000 Blackwell GPU with 96~GB of memory, an Intel Xeon w5-2455X CPU and 256~GB of DDR5 system memory. The software environment used Ubuntu 22.04, Python 3.14.7 and PyTorch 2.12.0 with CUDA 13.0. Training used bfloat16 mixed precision on a single GPU. The main comparisons used five random seeds (417, 725, 3921, 4927 and 7202). Deterministic CUDA execution was enabled for reproducibility. ACSD training required approximately 540~s per epoch. Inference on the \(248^3\)-voxel low-resolution synchrotron CT volume, corresponding to a physical extent of approximately \(0.652\times0.652\times0.652~\mathrm{mm}^3\), required approximately 1160~s. The resulting \(5952\times5952\times248\) reconstructed mask stack was resampled to a \(1488^3\) isotropic grid for geometric analysis and fiber tracking. This resampling provided a common spatial grid for analysis without introducing additional through-thickness information. Cylindrical template matching and fiber tracking required approximately 2~h on the same workstation.

\begin{figure}[ht]
 \centering
 \includegraphics[width=0.9\textwidth]{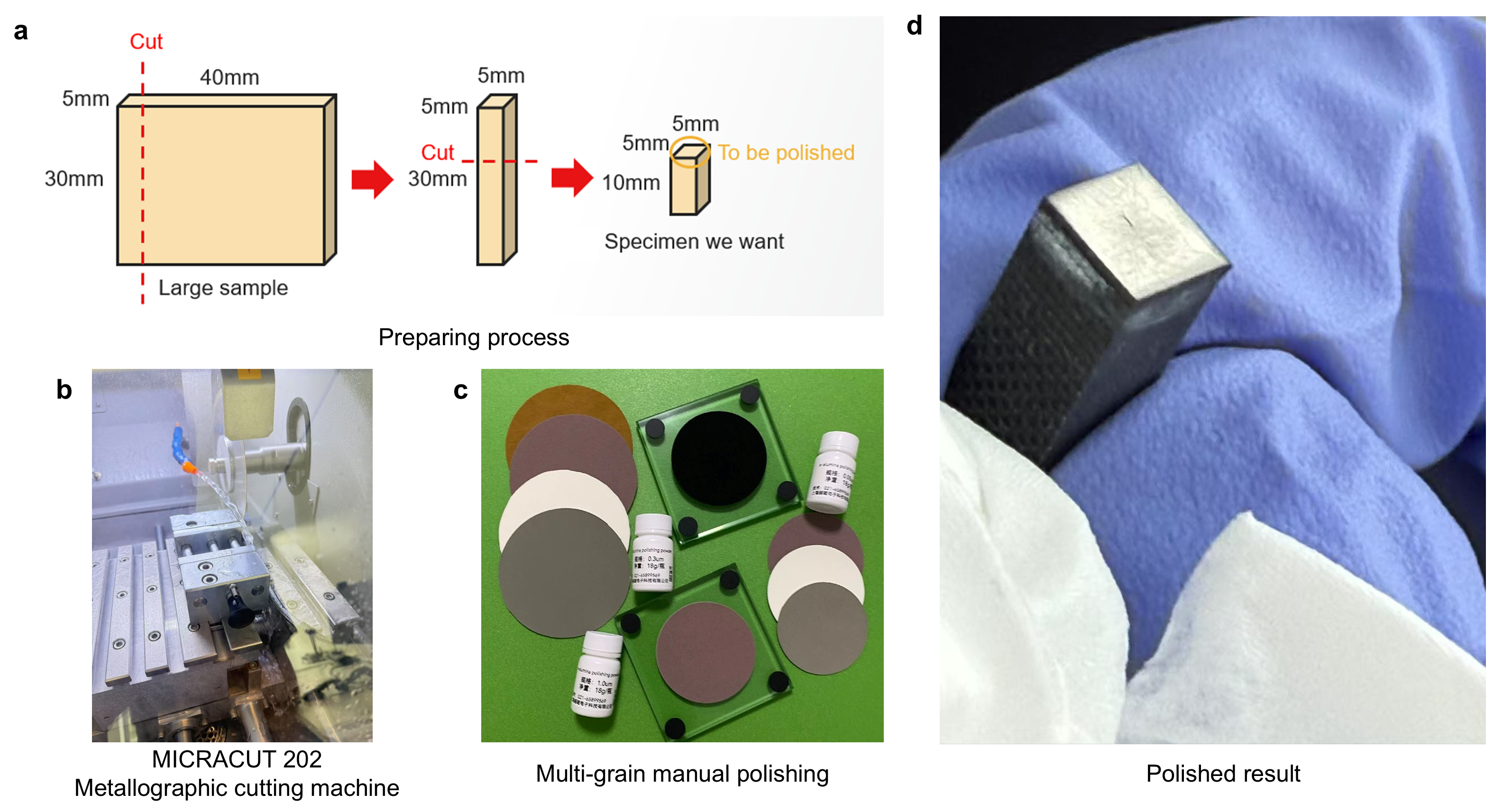}
 \caption{\textbf{Specimen preparation for multimodal imaging.} \textbf{a}, Schematic showing the sectioning of a \(40 \times 30 \times 5\) mm CFRP sample into \(5 \times 5 \times 10\) mm specimens. \textbf{b}, Metallographic sectioning using a MICRACUT 202 cutting machine. \textbf{c}, Materials used for multi-grain manual polishing, including polishing cloths and 1.0 and 0.3 \(\mu\)m \(\alpha\)-alumina powders and 0.05 \(\mu\)m \(\gamma\)-alumina powder. \textbf{d}, Mirror-finished specimen end face after polishing, prepared for subsequent microscopic imaging.}
 \label{SI:sample}
\end{figure}

\begin{figure}[ht]
 \centering
 \includegraphics[width=\textwidth]{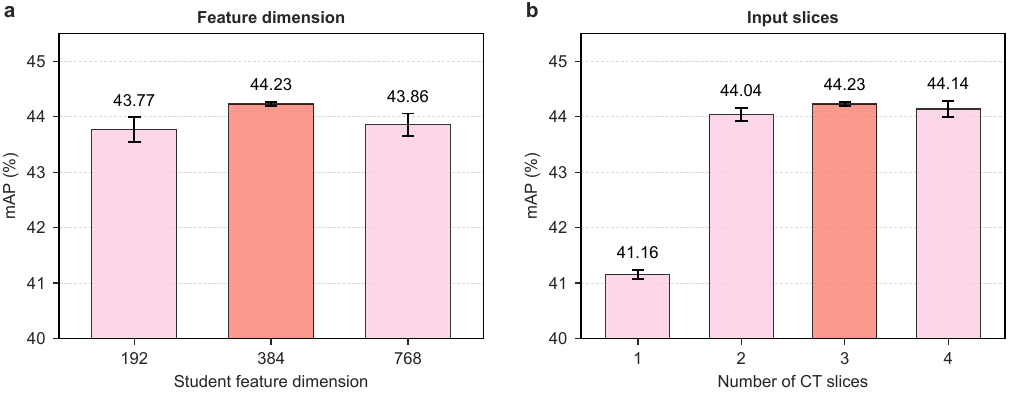}
 \caption{\textbf{Ablation of student feature dimensionality and multi-slice CT input.} Results are reported as mean $\pm$ standard deviation across three random seeds. \textbf{a}, mAP with student feature dimensions of 192, 384 and 768, with the teacher dimension fixed at 768. Increasing the student dimension from 192 to 384 improves performance, while further expansion to 768 provides no additional gain, supporting the use of a compact 384-dimensional student representation. \textbf{b}, mAP using one to four adjacent CT slices as input. Multi-slice context substantially improves performance over the single-slice setting, with performance remaining relatively stable from two to four slices. Three slices achieve the highest mean mAP and are used in the final model.}
 \label{SI:ablation}
\end{figure}

\clearpage

\subsection*{Note 2: Micro-CT surface extraction and cross-modal image registration}

To establish spatial correspondence between SEM images and micro-CT volumes, we first constructed a representative micro-CT image of the polished specimen end face. Because the specimen surface did not coincide exactly with a single reconstructed CT slice, its local depth was estimated from the first appearance of high-intensity specimen pixels along the depth direction. The resulting surface-depth map was processed using mode filtering to suppress local fluctuations. The filtered surface indices were then used to construct a representative CT end-face image for registration.

Before registration, SEM images were denoised using non-local means filtering. An initial affine registration was performed between the full SEM and CT end-face images to establish specimen-level alignment. The CT image was then divided into \(48\times48\)-pixel tiles directly on its native sampling grid, preserving the original voxel size of \(2.630~\mu\mathrm{m}\). The SEM images were resampled to a pixel size exactly one twenty-fourth of the native CT voxel size, corresponding to approximately \(0.110~\mu\mathrm{m\,px^{-1}}\). For each CT tile, the initial affine transformation was used to locate the corresponding SEM region, from which a larger \(1408\times1408\)-pixel patch was extracted to provide margins for local alignment. 
Each CT tile and its corresponding SEM patch were then registered independently using rigid registration to correct residual local misalignment. Following refinement, the aligned SEM region was cropped to \(1152\times1152\) pixels to obtain the final registered SEM–CT pair. Examples of the resulting alignment are shown in Supplementary Fig.~\ref{SI:reg}.

\begin{figure}[ht]
 \centering
 \includegraphics[width=0.80\textwidth]{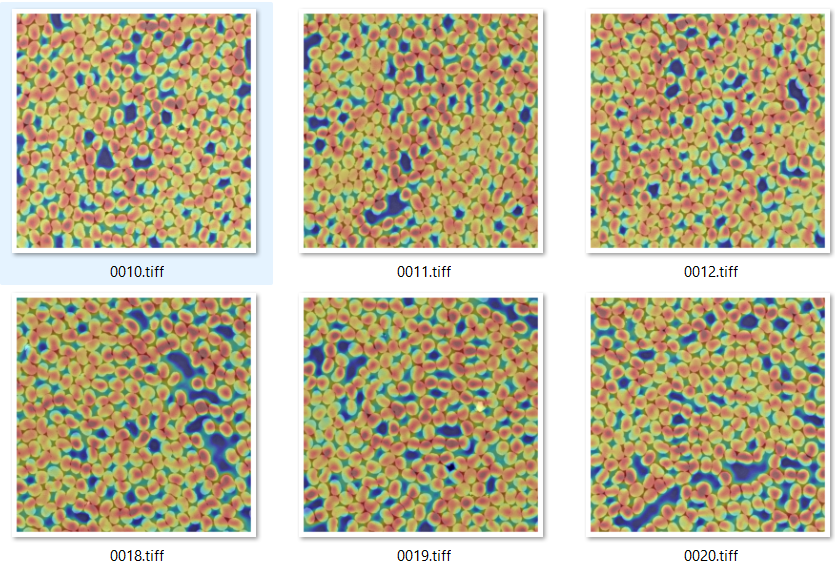}
 \caption{\textbf{Visualization of fine registration between SEM and micro-CT images.} Registration results are shown for six pairs. Grayscale images show the SEM observations, with the interpolated micro-CT images overlaid as heat maps. Low- and high-density features in the micro-CT images spatially coincide with the corresponding structures in the SEM images across all specimens, demonstrating consistent cross-modal alignment after fine registration.}
 \label{SI:reg}
\end{figure}

\subsection*{Note 3: Synchrotron X-ray CT data preparation for independent volumetric validation}

An independent synchrotron X-ray CT dataset previously reported \cite{mageira2026comprehensive} was used for volumetric validation. The dataset was acquired at the DanMAX beamline of the MAX IV synchrotron facility with a voxel size of approximately \(0.55~\mu\mathrm{m}\). The synchrotron volume served as the high-resolution reference (HR reference).

\begin{figure}[ht]
 \centering
 \includegraphics[width=0.80\textwidth]{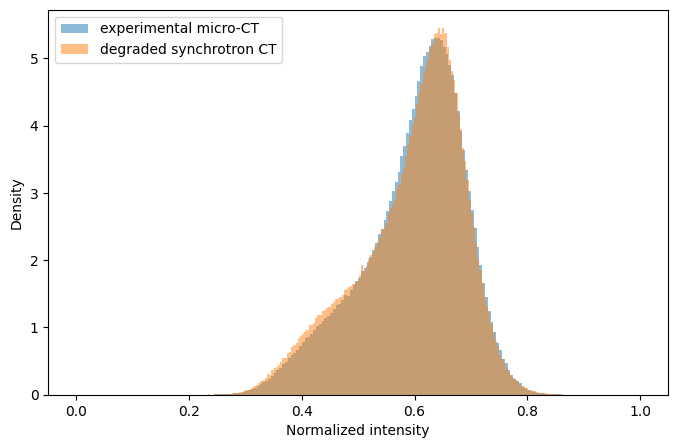}
 \caption{\textbf{Intensity-distribution comparison for synchrotron-to-micro-CT downsampling calibration.} Intensity histograms of the downsampled synchrotron X-ray CT volume and experimental micro-CT data after percentile-based rescaling. The Gaussian blur strength was calibrated to align the dominant intensity-distribution peaks of the two datasets.}
 \label{SI:peak}
\end{figure}

A low-resolution volume was generated from the HR reference following established downsampling procedures \cite{cardoso2015scale,diaz2017downsampling,sjolin2017compression}. The synchrotron volume was blurred using \texttt{scipy.ndimage.gaussian\_filter} with \(\sigma=3.0\) before downsampling to the micro-CT voxel size of \(2.630~\mu\mathrm{m}\). The blur strength was calibrated by aligning the dominant intensity-distribution peak of the downsampled synchrotron data with that of the experimental micro-CT after percentile-based rescaling using the 0.1st and 99.9th percentiles. The resulting intensity distributions are compared in Fig.~\ref{SI:peak}.

A \(248^3\)-voxel region was selected from the resulting low-resolution volume, covering approximately \(0.652\times0.652\times0.652~\mathrm{mm}^3\). This volume served as the LR input to the trained micro-CT student. The same LR input was upsampled using trilinear interpolation to construct the interpolated LR baseline for comparison with the HR reference.

\subsection*{Note 4: Adaptation of cross-modal knowledge distillation methods}

We adapted four feature-based knowledge distillation methods for the comparison in Table \ref{tab:main_results}. All methods used the same student architecture and supervised segmentation loss, with an additional distillation objective applied to the fused features before decoder upsampling. Let \(F_T\) and \(F_S\) denote the teacher and student fusion features, respectively. The teacher was kept fixed, and its features were precomputed for student training. Each method was optimized using
\[
\mathcal L=\mathcal L_{\mathrm{seg}}+\lambda_m\mathcal L_m,
\]
where \(\mathcal L_{\mathrm{seg}}\) combines flow regression and foreground classification losses, \(\mathcal L_m\) is the method-specific distillation loss, and \(\lambda_m\) controls its contribution.

\subsubsection*{Feature alignment}

Feature alignment transfers the teacher's intermediate representation through direct feature matching. A learnable \(1\times1\) convolution \(P\) projects the student fusion features to the teacher channel dimension. The projected student features are aligned with the corresponding teacher features using mean squared error,
\[
\mathcal L_{\mathrm{FA}}
=\operatorname{MSE}\left(P(F_S),F_T\right).
\]
The loss enforces elementwise correspondence between the two representations across all spatial locations and channels.

\subsubsection*{Frequency decomposition}

Frequency decomposition aligns the low-frequency components of the teacher and student features. The student features are first projected to the teacher channel dimension using a learnable \(1\times1\) convolution \(P\). A 2D Fourier transform is then applied independently to each feature channel. The low-frequency representation is obtained by applying a circular frequency mask and transforming the filtered spectrum back to the spatial domain,
\[
\mathcal H_{\rho}(F)
=\operatorname{Re}\left[
\mathcal F^{-1}
\left(M_{\rho}\odot\mathcal F(F)\right)
\right],
\]
where \(\mathcal F\) denotes the centered 2D Fourier transform and \(M_{\rho}\) retains frequencies within a radius of \(\rho\min(H,W)/2\) from the spectral center. The distillation loss is defined as
\[
\mathcal L_{\mathrm{FD}}
=\operatorname{MSE}\left(
\mathcal H_{\rho}(P(F_S)),
\mathcal H_{\rho}(F_T)
\right).
\]
The cutoff ratio was set to \(\rho=0.25\), restricting feature alignment to the retained low-frequency components.

\subsubsection*{Shared-private decomposition}

Shared-private decomposition separates each modality's features into shared and private representations. The student features are first mapped to the teacher channel dimension by a learnable \(1\times1\) adapter \(P\). A shared encoder \(E_{\mathrm{sh}}\) is applied to both modalities, while modality-specific encoders \(E_S\) and \(E_T\) extract their private components,
\[
\begin{aligned}
X_S &= P(F_S), & X_T &= F_T,\\
G_m &= E_{\mathrm{sh}}(X_m),&
U_m &= E_m(X_m),\quad m\in\{S,T\}.
\end{aligned}
\]
The shared representations are aligned using mean squared error,
\[
\mathcal L_{\mathrm{shared}}
=\operatorname{MSE}(G_S,G_T).
\]
An orthogonality constraint reduces correlations between the shared and private components. Let \(\widetilde G_m\) and \(\widetilde U_m\) denote the corresponding features flattened along spatial dimensions and normalized independently for each channel. The orthogonality loss is
\[
\mathcal L_{\mathrm{orth}}
=\frac{1}{2}\sum_{m\in\{S,T\}}
\operatorname{Mean}\left[
\left(
\widetilde G_m\widetilde U_m^\top
\right)^{\odot 2}
\right],
\]
where \(\odot 2\) denotes elementwise squaring. Separate reconstruction layers \(R_S\) and \(R_T\) recover the input representations from the concatenated shared and private components,
\[
\mathcal L_{\mathrm{rec}}
=\frac{1}{2}\sum_{m\in\{S,T\}}
\operatorname{MSE}
\left(R_m([G_m,U_m]),X_m\right).
\]
The resulting distillation objective is
\[
\mathcal L_{\mathrm{SPD}}
=\mathcal L_{\mathrm{shared}}
+\mathcal L_{\mathrm{orth}}
+\mathcal L_{\mathrm{rec}}.
\]

\subsubsection*{Cross-affinity distillation}

Cross-affinity distillation transfers spatial feature relationships using cross-modal affinity matrices. The teacher and student fusion features have a spatial resolution of \(72\times72\). The student features are projected to the teacher channel dimension using a learnable \(1\times1\) convolution. Both representations are then adaptively average-pooled to three spatial resolutions \(s\in\{36,18,9\}\). After spatial flattening and channel-wise \(\ell_2\) normalization, the resulting features are denoted by \(S_s\) and \(T_s\).

At each resolution, the cross-affinity matrix and student self-affinity matrix are calculated as
\[
A_s^{\mathrm{cross}}=T_s^\top S_s,
\qquad
A_s^{\mathrm{self}}=S_s^\top S_s.
\]
The distillation objective minimizes the mean absolute difference between the two affinity matrices across the three resolutions,
\[
\mathcal L_{\mathrm{CAD}}
=\frac{1}{3}\sum_{s\in\{36,18,9\}}
\operatorname{MAE}\left(
A_s^{\mathrm{cross}},
A_s^{\mathrm{self}}
\right),
\]
where \(\operatorname{MAE}\) denotes the mean absolute error over all affinity matrix elements and samples. Teacher features were cached at a resolution of \(36\times36\) and further pooled for the smaller scales.

\subsection*{Note 5: Volumetric structural characterization and fiber tracking}

\subsubsection*{Structural characterization}

Structural statistics were evaluated within the same physical ROI for the HR reference, interpolated LR and PCRD reconstruction. Fiber volume fraction was determined using the entropy-based histogram thresholding method of Kapur et al. \cite{kapur1985new} for the HR reference at its native resolution and for the interpolated LR on the \(1488^3\) grid. The method selects the threshold by maximizing the combined entropies of the two histogram classes. For the PCRD reconstruction, fiber volume fraction was calculated directly from the predicted binary masks.

Equivalent fiber diameter and nearest-neighbor distance require individual fiber instances. For the HR reference, these instances were obtained using Cellpose-SAM, followed by the similar filtering procedure used for SEM annotations to remove detections inconsistent with the expected fiber morphology. For the PCRD reconstruction, the fiber-instance masks were generated directly by the trained student from the LR input and required no additional instance segmentation. For each retained fiber cross-section with area \(A\), the equivalent diameter was calculated as
\[
d_{\mathrm{eq}}=2\sqrt{\frac{A}{\pi}}.
\]
Nearest-neighbor distance was calculated as the minimum Euclidean distance between the centroid of each retained fiber instance and the centroids of all other retained instances. Instances intersecting the ROI boundary were excluded from these measurements.

\subsubsection*{Cylindrical template matching}

Fiber centerlines were extracted from volumetric data using cylindrical template matching followed by bidirectional line tracing. The same procedure and tracking parameters were applied to the HR reference, interpolated LR and PCRD reconstruction on a common \(1488^3\) isotropic grid. The tracking voxel size was approximately \(0.4383~\mu\mathrm{m}\), corresponding to the \(652.24~\mu\mathrm{m}\) physical extent of the validation volume.

A bank of binary cylindrical templates was generated with a length of 80 voxels (\(35.07~\mu\mathrm{m}\)), an outer radius of 8.5 voxels (\(3.73~\mu\mathrm{m}\)) and a correlation-mask radius of 10 voxels (\(4.38~\mu\mathrm{m}\)). Template orientations were sampled at \(2^\circ\) intervals within a \(48^\circ\) cone around the fiber direction. To reduce computation, orientation candidates were retained using angular sparsification factors of 1, 2 and 8 for deviations of \(0^\circ\)–\(24^\circ\), \(24^\circ\)–\(36^\circ\) and \(36^\circ\)–\(48^\circ\), respectively.

For each template orientation, local normalized cross-correlation was calculated using FFT-based correlation. The highest correlation coefficient and its corresponding template orientation were retained at each voxel, producing a correlation field and a local orientation field for subsequent line tracing. Correlation was evaluated blockwise, with overlapping margins to preserve valid responses across block boundaries. Template correlation was performed in batches of one template.

\subsubsection*{Bidirectional fiber tracing}

Candidate tracing seeds were selected from voxels with a normalized correlation coefficient of at least 0.47 and processed in descending order of correlation strength. Each seed was traced in both directions along the locally estimated fiber orientation.

At each tracing step, candidate continuation points were searched within a cone of length 24 voxels (\(10.52~\mu\mathrm{m}\)) and total opening angle \(4^\circ\). The minimum step length was set to half the cone length, corresponding to 12 voxels (\(5.26~\mu\mathrm{m}\)). Candidate points were scored using their local correlation and the angular consistency between the current direction, the candidate orientation and the connecting displacement. The directional weighting coefficient was 0.20, and tracing continued only when the best continuation quality exceeded 0.18.

The forward and backward traces were combined into a single centerline. Traces shorter than 249.75 voxels (\(109.47~\mu\mathrm{m}\)) were rejected. Accepted traces were used to suppress nearby candidates within a minimum line separation of 11.25 voxels (\(4.93~\mu\mathrm{m}\)), reducing duplicate tracking of the same fiber. Tracing was performed on the CPU without imposing a maximum number of seeds or accepted lines.

\subsubsection*{Centerline matching and tracking metrics}

The extracted centerlines were transformed to a common physical coordinate system and represented on a \(5952^3\) coordinate grid, corresponding to a sampling interval of approximately \(0.1096~\mu\mathrm{m}\). Reference and reconstructed centerlines were interpolated onto common axial planes sampled every five grid positions.

At each sampled plane, transverse centerline positions were matched using mutual nearest neighbors. A match was accepted when the reference point and reconstructed point were each other's nearest neighbor and their transverse separation did not exceed \(3~\mu\mathrm{m}\). Matching was performed independently for the interpolated LR and PCRD reconstruction.

Reference centerline coverage was calculated from the lengths of segments connecting consecutive sampled reference points. A segment was considered covered only when both endpoints were successfully matched. The coverage was defined as
\[
C=
\frac{\sum_{e\in\mathcal E_{\mathrm{matched}}}L_e}
{\sum_{e\in\mathcal E_{\mathrm{ref}}}L_e}
\times100\%,
\]
where \(L_e\) is the three-dimensional length of a sampled reference centerline segment, \(\mathcal E_{\mathrm{ref}}\) contains all eligible sampled reference segments and \(\mathcal E_{\mathrm{matched}}\) contains segments with both endpoints matched.

Position deviation was defined as the transverse Euclidean distance between corresponding centerline points. Local tangent directions were estimated using the displacement between sampled centerline points five positions before and after each evaluated point, with the window truncated near line endpoints. Orientation deviation was calculated as
\[
\theta=
\cos^{-1}\left(
\left|\mathbf t_{\mathrm{ref}}\cdot\mathbf t_{\mathrm{pred}}\right|
\right),
\]
where \(\mathbf t_{\mathrm{ref}}\) and \(\mathbf t_{\mathrm{pred}}\) are unit tangent vectors of matched reference and reconstructed centerlines. The absolute dot product accounts for the direction ambiguity of fiber centerlines.

For pointwise statistics, each sampled reference point was assigned a length weight equal to half the sum of its adjacent segment lengths, with endpoint weights based on the single adjacent segment. Each reconstruction was evaluated on its own matched reference support.

\end{document}